\documentclass[10pt,twocolumn,letterpaper]{article}

\usepackage{iccv}
\usepackage{times}
\usepackage{epsfig}
\usepackage{graphicx}
\usepackage{amsmath}
\usepackage{amssymb}
\usepackage{multirow}
\usepackage{booktabs}
\usepackage{graphicx} 
\usepackage[export]{adjustbox}
\usepackage[dvipsnames]{xcolor}{\tiny }
\usepackage{xcolor,colortbl}
\usepackage{color, colortbl}
\usepackage{xcolor}

\usepackage{sidecap}
\usepackage{pgfplots} 

\usepackage{float}
\definecolor{purple}{rgb}{0.65,0,0.65}
\definecolor{dark_green}{rgb}{0, 0.5, 0}
\definecolor{blueish}{rgb}{0.0, 0.3, .6}
\definecolor{lightgreen}{RGB}{238, 252, 241}
\definecolor{lightred}{RGB}{231, 187, 187}
\definecolor{darkred}{RGB}{198, 129, 129}

\definecolor{tabhighlight}{HTML}{e5e5e5}

\usepackage{graphicx, amsmath, amssymb, caption, subcaption, multirow, overpic}
\definecolor{tabhighlight}{HTML}{e5e5e5}
\definecolor{citecolor}{HTML}{0071bc}

\usepackage[pagebackref=true,breaklinks=true,letterpaper=true,colorlinks,bookmarks=false]{hyperref}

\iccvfinalcopy 

\ificcvfinal\fi

\begin{document}

\title{Diffusion Based Augmentation for Captioning and Retrieval in Cultural Heritage}

\author{Dario Cioni$^1$ \qquad Lorenzo Berlincioni$^1$ \qquad Federico Becattini$^2$ \qquad Alberto del Bimbo$^1$\\
$^1$MICC, University of Florence\qquad$^2$University of Siena\\
{\tt\small dario.cioni@stud.unifi.it}\quad{\tt\small \{lorenzo.berlincioni,alberto.delbimbo\}@unifi.it}\\ {\tt\small federico.becattini@unisi.it}
}
\maketitle
\ificcvfinal\thispagestyle{empty}\fi

\begin{abstract}

Cultural heritage applications and advanced machine learning models are creating a fruitful synergy to provide effective and accessible ways of interacting with artworks. Smart audio-guides, personalized art-related content and gamification approaches are just a few examples of how technology can be exploited to provide additional value to artists or exhibitions.
Nonetheless, from a machine learning point of view, the amount of available artistic data is often not enough to train effective models. Off-the-shelf computer vision modules can still be exploited to some extent, yet a severe domain shift is present between art images and standard natural image datasets used to train such models. As a result, this can lead to degraded performance.
This paper introduces a novel approach to address the challenges of limited annotated data and domain shifts in the cultural heritage domain. By leveraging generative vision-language models, we augment art datasets by generating diverse variations of artworks conditioned on their captions. This augmentation strategy enhances dataset diversity, bridging the gap between natural images and artworks, and improving the alignment of visual cues with knowledge from general-purpose datasets. The generated variations assist in training vision and language models with a deeper understanding of artistic characteristics and that are able to generate better captions with appropriate jargon. 
\end{abstract}

\begin{figure*}[t]
	\centering
        \includegraphics[width=0.96\textwidth]{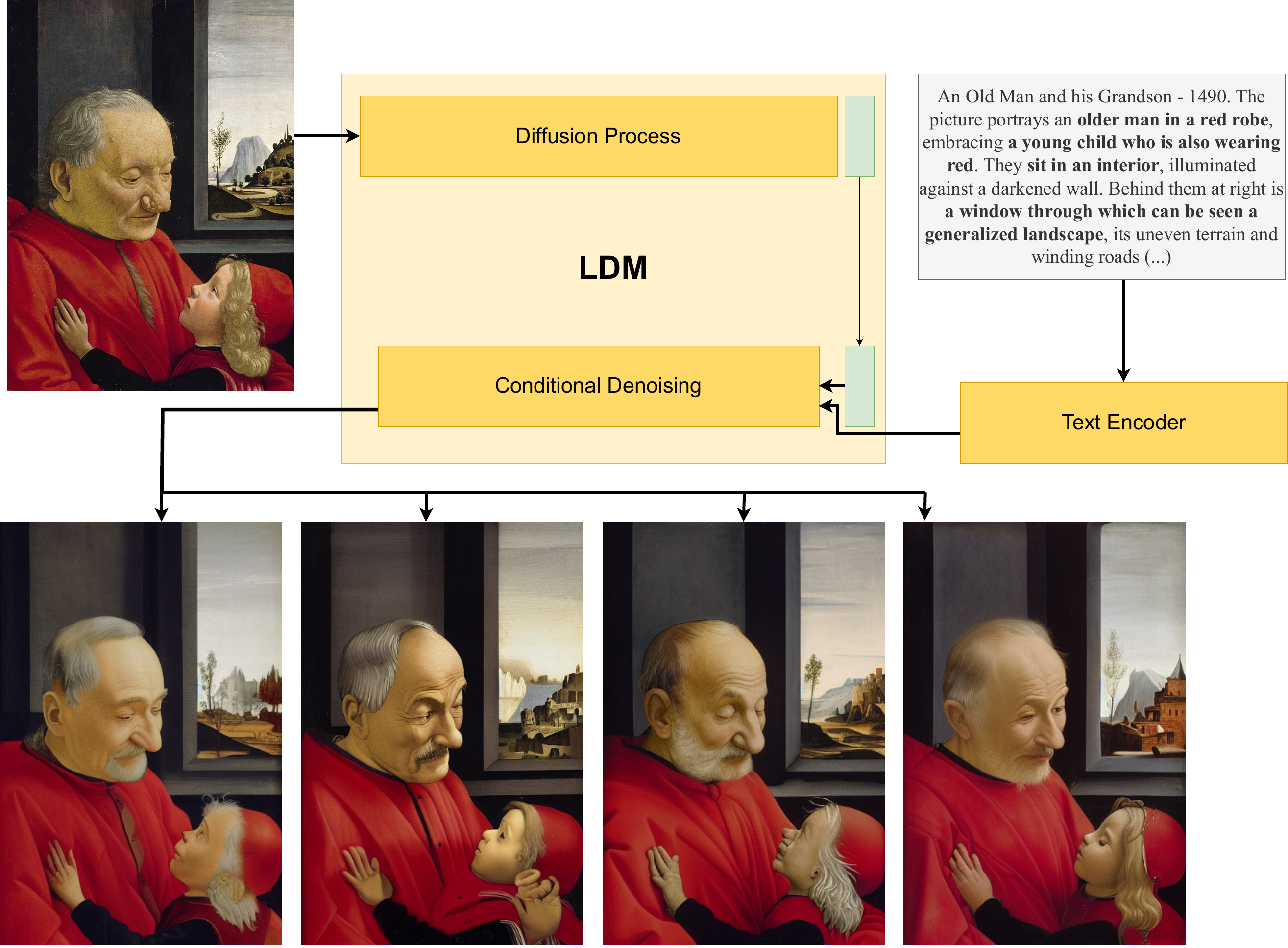}
 \caption{Schematic illustrating the data augmentation pipeline. The conditional generative model allows for both and image and text input while in its \textit{Image\&Text}$\longrightarrow$\textit{Image} configuration. We provide the model both the original artwork along with its detailed textual analysis from \cite{stefanini2019artpedia} and use the diffusion model's outputs as new datapoints for training other models for downstream tasks.}
	\label{img:schematic}
\end{figure*}

\section{Introduction}
\label{sec:intro}
Deep learning applications on fine art suffer from an obvious scarcity of data, due to the fact that an artwork is usually a unique piece. In addition, tasks involving both vision and language require on the one hand the modeling of technical language with domain-specific jargon and on the other hand the understanding of difficult and underrepresented visual concepts, such as abstract or stylized drawings. These difficulties entail a challenge for learning algorithms, which would benefit from a large collection of annotated data.
A simple solution is to leverage models pre-trained on general-purpose datasets to address relevant tasks for cultural heritage, such as retrieval, visual question answering or captioning. However, the effect is that such models tend to underperform when applied in the cultural heritage domain. In fact, when the scope moves from real-world images to paintings and other more abstract representations, there is a strong domain shift compared to standard training data, being it composed of natural images.
A standard approach to deal with data scarcity is to leverage data augmentation, slightly perturbing the training data to improve variability and let the trained model generalize better. In the vision domain, perturbations usually include adding noise, altering pixel values or changing the overall orientation or illumination of the scene. We argue that these augmentations may indeed alter the semantics of the artworks, where color and spatial distribution of objects can convey significant meanings that are necessary to interpret the painting.

In this paper, we address the above-mentioned limitations by proposing a data augmentation strategy for paintings that has the twofold advantage of increasing the training data as well as preserving the content.
In particular, we explore the benefits of augmenting artwork datasets for image captioning. To this end, we leverage both textual descriptions of the paintings and a diffusion model to create several variations of the artworks. By conditioning the diffusion model on the captions, we generate a variability in the visual domain that aids the grounding of objects and entities expressed in artistic form with the technical jargon that describes them.
What we propose is therefore an image augmentation at a semantic level, generating multiple variations of artworks while retaining their content and style.
By leveraging the expert knowledge of art critics contained in painting descriptions and the natural language understanding capabilities of state-of-the-art generative models, we aim to provide a sophisticated augmentation pipeline capable of generating a sufficient intra-class variability of depicted concepts to enable an effective learning (Fig. \ref{img:schematic}).

Our main contributions presented in this paper are
\begin{itemize}
    \item We propose a data augmentation technique for low-data regime cultural heritage tasks, that works at a semantic level rather than at a pixel-intensity level as standard data augmentations in vision.
    \item Thanks to our data augmentation strategy based on diffusion models we can favor a visual grounding of linguistic concepts, which in the cultural heritage domain are often expressed using technical and domain-specific jargon.
    \item We show the benefits of the proposed augmentation strategy in captioning tasks as well as cross-domain retrieval tasks.
\end{itemize}

\section{Related Works}
\paragraph{Computer Vision for Cultural Heritage}
In the domain of cultural heritage, several computer vision approaches have been proposed in the literature. Artwork classification~\cite{mensink2014rijksmuseum,tan2016ceci,del2019noisyart,cetinic2018fine,milani2021dataset} and recognition~\cite{del2019webly,temmermans2011mobile,jin2017artwork} have often been placed at the center of such approaches, sometimes with the end-goal to develop user-engagement applications~\cite{becattini2016imaging, majd2017impact, aruanno2022tintoretto, bongini2022gpt}.
In this paper, we mostly deal with the task of image captioning, which implies the automatic generation of a natural language textual description of an image based only on the visual input. This has been an extensively addressed research topic in recent years \cite{stefanini2022show, wang2022git, li2022blip}, but not many contributions have been made in the domain of art historical data. In this particular domain, which shifts from the one of natural images, the complexity of the task increases due to a simultaneous lack of labeled data and an increased abstraction.

Currently, available painting datasets with descriptions are constructed by downloading descriptions from online museums or annotating descriptions by crowdsourcing.
The Artpedia \cite{stefanini2019artpedia} dataset is composed of paintings paired with textual descriptions from WikiPedia. The dataset thus provides information about artworks and their context and each sentence is categorized as either a visual sentence or a contextual sentence. Visual sentences describe the visual content of the painting, while contextual sentences provide information that cannot be inferred from raw pixels alone. Such information includes, for instance, the name of the painter, its artistic style, or the museum in which it is kept.
The dataset was originally introduced as a dataset for cross-modal retrieval as well as captioning and it has been further annotated for visual question answering purposes in \cite{bongini2020visual}.
Similarly, the AQUA dataset~\cite{garcia2020dataset} has been proposed to train visual question answering models in the cultural heritage domain.
More recently, ArtCap \cite{artcap} provided an image captioning dataset containing 3,606 paintings, each one associated to five textual descriptions, with a mean length for each caption of 11 words.

A larger example of an artwork dataset is presented in \cite{del2019webly} consisting of more than 80K webly-supervised images from 3120 classes, and a subset of 200 classes with more than 1300 verified images. Text and metadata for each class is also provided, to support zero-shot learning and other multi-modality techniques in general.
An ontological knowledge base has been exploited in \cite{becattini2023viscounth} to create a large-scale cultural heritage dataset, annotated with visual and contextual data. The authors adopted ArCo \cite{carriero2019arco}, the Italian cultural heritage knowledge graph, to extract information about approximately 500K cultural assets and leveraged a semi-automatic annotation approach for generating 6.5M question-answer pairs.

\paragraph{Generative Models for Data Augmentation}
Data augmentation can be defined as the process through which data can be transformed without changing its semantics. By using this definition we can tie the efficacy of an augmentation method to a task and not to the type of data alone. In most of the computer vision tasks that work with natural images, the usual augmentation strategies involve flipping the image, adding random noise, and changing its brightness and colors. When it comes to fine-art tough, such changes might be detrimental due to the strict relation between the semantics of the original art piece and its details
(i.e. the relative position of characters in religious art, the use of strong light contrast in a Caravaggio painting, or the symbolic choice of a particular color).
An attempt to augment training data for object detection in artworks has been recently proposed~\cite{jeon2020object}, where a style transfer model is applied to natural images to generate images that resemble paintings.
A possible approach to obtain a larger, more diverse, dataset has been explored in many works outside of the scope of cultural heritage. In these cases, the input data is used to train a generative model, which in turn will produce new data coming from the training domain.
\cite{shrivastava2017learning,BARTH2020105378,augGAN,Hoffman2018CyCADACA, berlincioni2021multiple}.
Also in~\cite{devaguptapu2019borrow} the authors used a CycleGAN~\cite{CycleGAN2017} for image-to-image translation of
thermal to pseudo-RGB data. The use of these frameworks to perform data
augmentation in order to improve the performance of a separate classifier has
been studied in multiple previous works such as
\cite{antoniuo-2018} in which they focus on improving one-shot
learning, and \cite{Bowles2018GANAA}, where segmentation of medical images is
enhanced by GAN augmented data. In \cite{pan2017virtual} synthetic data coming from a simulator is adapted and used to train an RL agent for autonomous driving.

Recently Diffusion Models (DM)~\cite{ddpm,ldm} reached new impressive levels, compared to GANs, both in terms of output quality and fidelity to the conditional inputs such as text or additional images and have been employed for data augmentation objectives such in \cite{trabucco, he2022synthetic}. Most of these applications focused mostly on evaluating the ability of diffusion models to generate synthetic data for classification problems, we instead are going to focus on different downstream tasks.
Latent Diffusion Models (LDM)~\cite{ldm} perform the diffusion process in a latent space learned by a convolutional auto-encoder. This allows to greatly reduce the training and inference cost of the model compared to pixel-based DMs, while maintaining a high visual fidelity. By introducing cross-attention layers in the diffusion model architecture, the generation can be conditioned on a wide variety of sources, including text and images. The popular Stable Diffusion model is based on LDM \cite{ldm}, bringing further improvements thanks to an internet-scale training.

Motivated by the recent success of large generative models, we posed the research question regarding whether diffusion models can be used to augment visual recognition datasets with synthetic images, especially when working in underrepresented domains such as cultural heritage. Our findings show that using images generated by a diffusion model, conditioned by a textual description, leads to improved performance compared to vanilla training as well as training using standard computer vision data augmentation techniques.

\begin{figure}[t]
    \centering
    \subfloat[\centering Artpedia]{{\includegraphics[width=0.49\columnwidth]{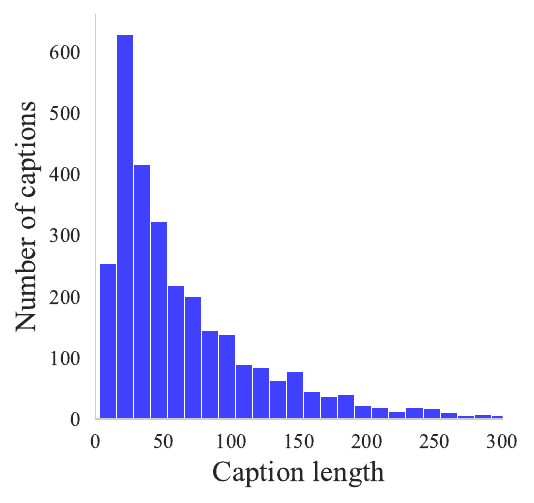} }}%
    \subfloat[\centering ArtCap]{{\includegraphics[width=0.49\columnwidth]{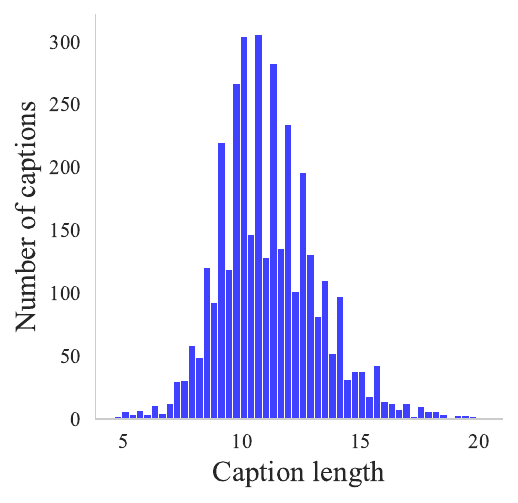} }}%
    \caption{Distribution of caption lengths in the Artpedia \cite{stefanini2019artpedia} and ArtCap \cite{artcap} datasets}%
    \label{img:caption_lenghts}%
\end{figure}

\section{Data}
Experiments were performed on both the Artpedia \cite{stefanini2019artpedia} and the ArtCap \cite{artcap} datasets. Albeit similar in structure these two datasets differ in multiple ways from one another. The \cite{artcap} dataset contains a fixed set of 5 sentences per artwork while \cite{stefanini2019artpedia} has on average, 3.1 visual sentences and 6.5 contextual sentences per artwork.
Artpedia contains a collection of 2,930 painting images, each associated with a variable number of textual descriptions, which are combined together into a single description. 
Overall, the dataset contains a total of 28,212 sentences. Out of these, 9,173 are labeled as visual sentences and the remaining 19,039 are categorized as contextual. In our work, we only consider visual sentences since we focus on augmenting images.

The respective syntactic style is also quite different in the two datasets: where Artpedia chooses paragraph-long academic descriptions, Artcap limits itself to shorter and simpler captions.

\begin{figure}[t]
  \centering
  \begin{tikzpicture}
    \node (image) at (0,0) {\includegraphics[width=0.3\columnwidth]{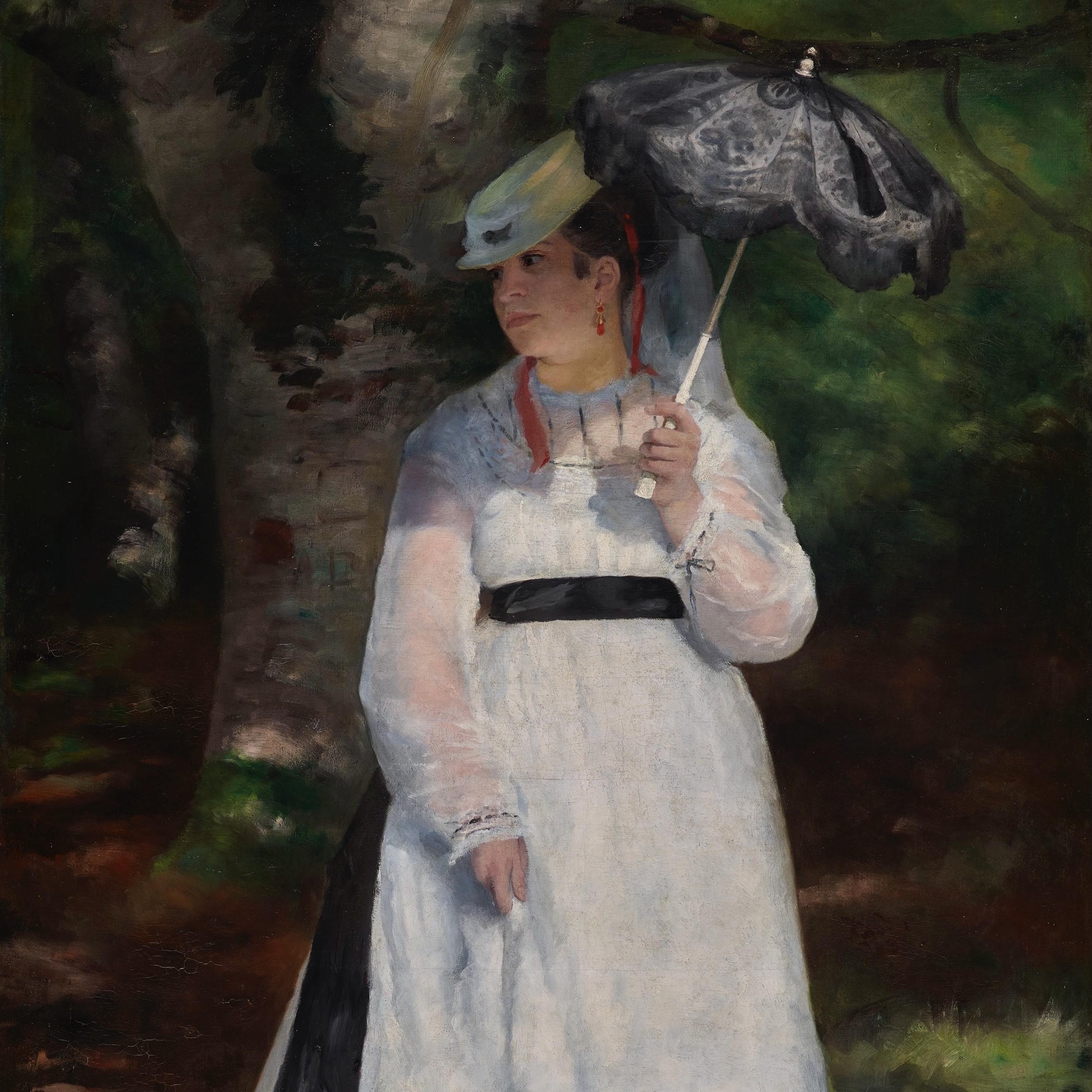}};

    \node [draw, fill=white, text width=0.65\columnwidth, align=left, right] (description) at (image.east) { \scriptsize
      The full-length painting depicts model Lise Tréhot posing in a forest. She wears a white muslin dress and holds a black lace parasol to shade her from the sunlight, which filters down through the leaves, contrasting her face in the shadow and her body in the light, highlighting her dress rather than her face. Lise is a full-length, almost life-size portrait of a young woman, standing in a forest clearing. She wears a small, pork pie straw hat with red ribbons, and a long white muslin dress with a long black sash; the dress is modestly buttoned to the neck and has long sheer sleeves.

    };
  \end{tikzpicture}
  \begin{tikzpicture}
    \node (image) at (0,0) {\includegraphics[width=0.3\columnwidth]{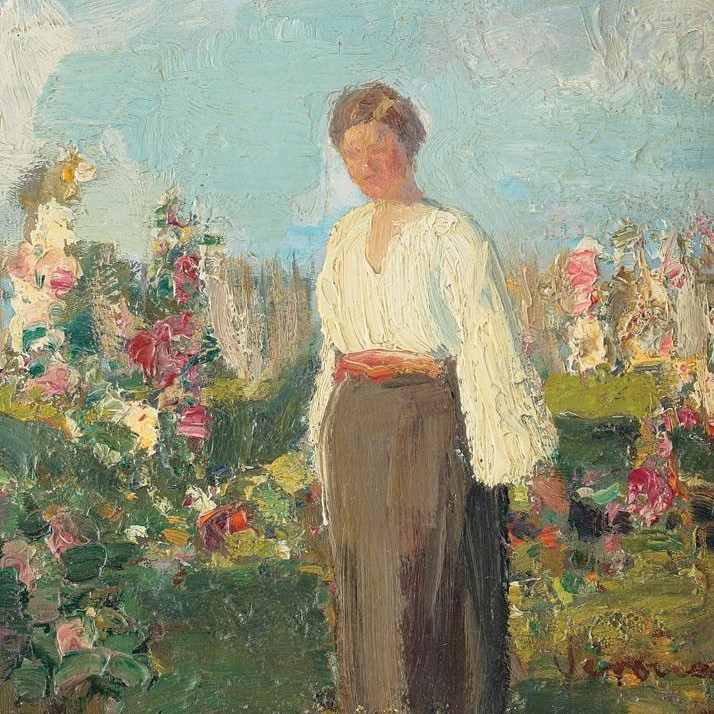}};

    \node [draw, fill=white, text width=0.65\columnwidth, align=left, right] (description) at (image.east) {
    \scriptsize
    The woman is standing in the flower garden wearing a long skirt;\\
    A woman is posing in front of the colorful garden;\\
    A woman in a long skirt in a flower garden on a spring or summer day;\\
    A woman is starring up in the sky in the garden;\\
    A woman is standing in the garden in a white shirt and brown skirt.\\
    };
  \end{tikzpicture}
  \caption{Samples of images along with their textual descriptions from Artpedia (top) and ArtCap (bottom) datasets.}
  \label{fig:dataset_samples}
\end{figure}

The word count distribution of the captions for the two datasets is shown in Fig.~\ref{img:caption_lenghts}. Note that, on average, a single visual sentence of Artpedia is composed by 22 words, and the caption is 70 words, which is considerably longer than most common Image Captioning datasets \cite{coco-captions,conceptual-captions,textcaps}. We also evaluated randomly sampling one of the visual sentences, but since each visual sentence describes only a small portion of the image, it led to worse results.
Both Artpedia and ArtCap provide validation and test splits, composed of 10\% and 10\% validation samples, and 10\% and 50\% test samples, respectively.
Samples from the two datasets are shown in Fig. \ref{fig:dataset_samples}.

\begin{figure}[h]
	\centering
        \includegraphics[width=0.98\columnwidth]{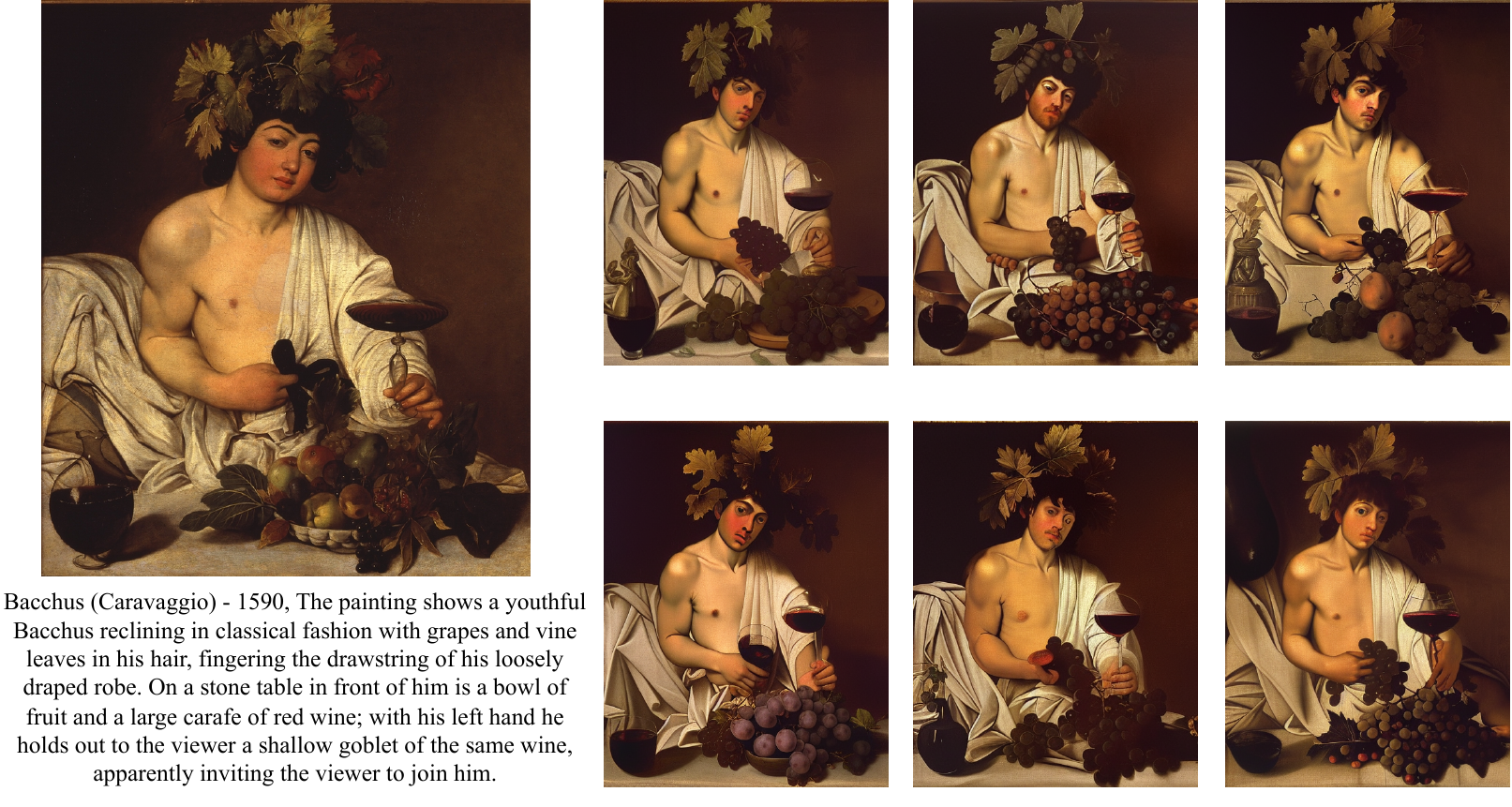}


	\centering
        \includegraphics[width=0.98\columnwidth]{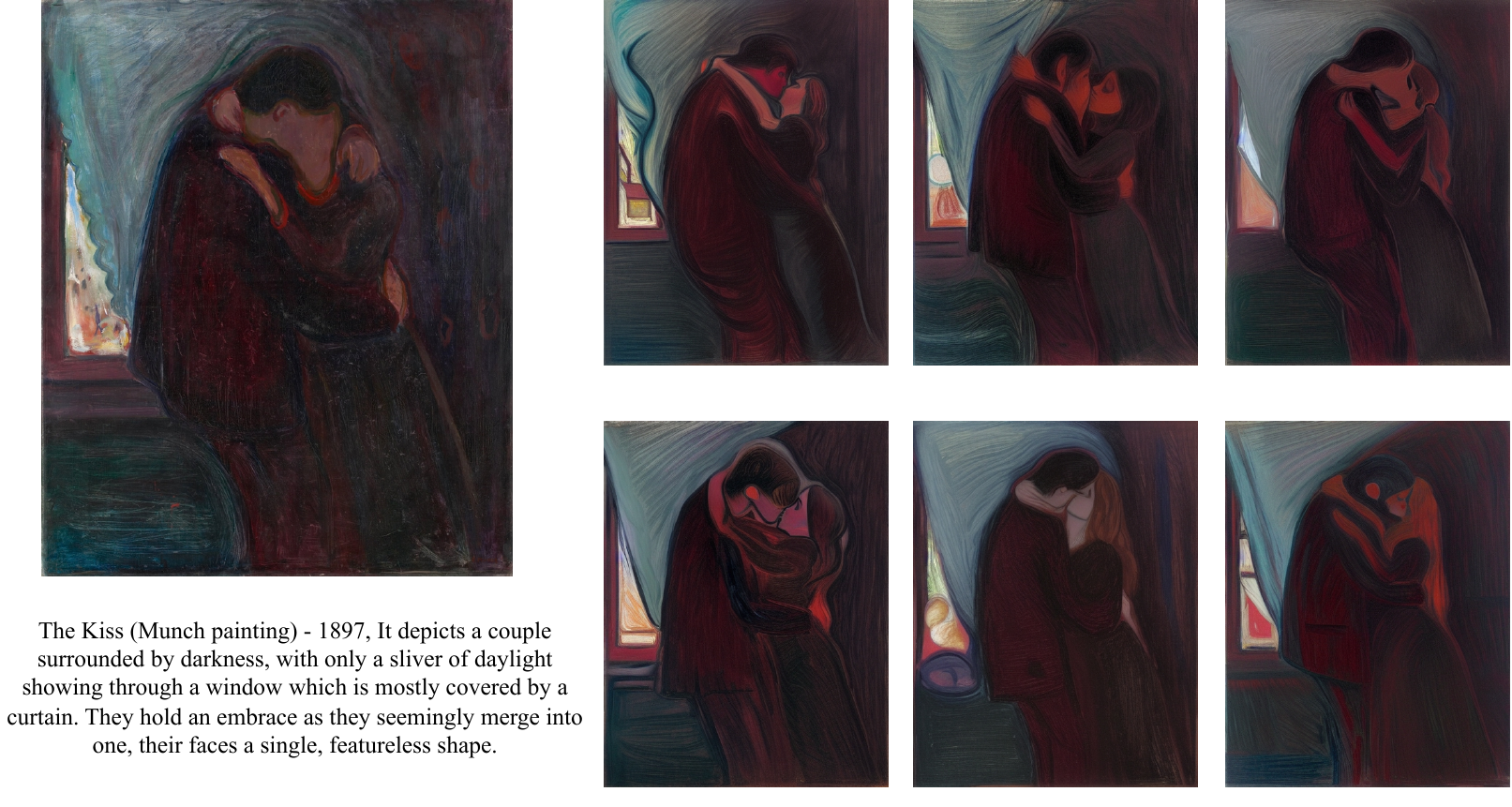}
 \caption{Samples of the augmented images. \textit{Left:} Original image and its caption.; \textit{Right:} Multiple samples of the augmented images using the combination of the provided description and the original input image.}
	\label{img:augmented}
\end{figure}

\section{Method}

In order to generate the augmented version of the datasets we employ a LDM (Latent Diffusion Model), Stable Diffusion\footnote{https://stability.ai/blog/stable-diffusion-public-release}, to generate multiple version of each image belonging to the original dataset as illustrated in Fig. \ref{img:schematic}.

In our work, we employed versions 1.4 and 1.5 of Stable Diffusion. In Stable Diffusion 1.4 the checkpoint was initialized with the weights of the Stable-Diffusion-v-1-2 checkpoint and subsequently fine-tuned on 225K steps at resolution 512x512 on the "laion-aesthetics v2 5+" subset of LAION dataset \cite{schuhmann2022laion} and 10\% dropping of the text-conditioning to improve classifier-free guidance sampling. In Stable Diffusion 1.5, the initialization checkpoint and the finetuning procedure is the same as Stable-Diffusion-1-4, but the finetuning is performed for more steps (595K).
We employed Stable Diffusion v1.4 to augment Artpedia \cite{stefanini2019artpedia} dataset and Stable Diffusion v1.5 for the ArtCap \cite{artcap} dataset. 

Given a dataset $\mathcal{D}$ of $N$ samples $(x_i, \mathbf{y}_i)$ formed by an image $x_i$ and a set of captions $\mathbf{y}_i$, we augment it by generating a set $S_i = \{(\tilde{x}_{i1},\mathbf{y}) \dots (\tilde{x}_{iM},\mathbf{y})\}$ of synthetic variations for each image $x_i$, obtaining a synthetic dataset $\tilde{\mathcal{D}}$ of $N\times M$ samples. Each variation was generated using both a textual prompt built by providing the caption and the original image to guide the generation. To obtain different images, the generation seed was changed for each variation (see Fig.~\ref{img:augmented}). 

\begin{figure}
    \centering
    \resizebox{0.99\columnwidth}{!}{%
    \begin{tikzpicture}
    \begin{axis}  
    [  
        ybar, 
        enlargelimits=0.15,
        legend style={at={(0.5,1)},anchor=north,legend columns=-1},     
        ylabel={Cosine similarity}, 
        symbolic x coords={(a), (b), (c)},  
        xtick=data,  
        nodes near coords,  
        nodes near coords align={vertical}, 
        bar width=20pt
        ]  
    \addplot coordinates {((a), 0.331) ((b), 0.333) ((c), 0.874)}; 
    \addplot coordinates {((a), 0.281) ((b), 0.302) ((c), 0.792)};
    \legend{Artpedia, ArtCap}  
    \end{axis}  
    \end{tikzpicture}  
    }
    \caption{Average cosine similarity between CLIP embeddings of: (a) real images and the associated captions; (b) synthetic images and the associated captions; (c) real images and their synthetic variations.}
    \label{fig:cosine-similarity}
\end{figure}
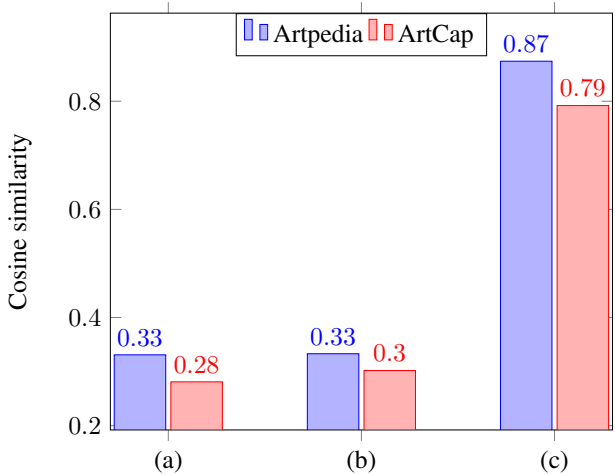

To gain an intuition of the quality of the synthetic dataset, we calculated an embedding of the text and images of each sample in $\mathcal{D}$ and $\tilde{\mathcal{D}}$ using a CLIP-ViT/B-16 model \cite{clip}. We can see from Fig.~\ref{fig:cosine-similarity} (a) and (b)  that the average cosine similarity between  images and their captions maintains a similar value in the original dataset $\mathcal{D}$ and the synthetic dataset $\tilde{\mathcal{D}}$, suggesting that synthetic images preserve the relation with the caption. 
Moreover, in (see Fig.~\ref{fig:cosine-similarity} (c)) we see that variations maintain a high similarity with the original images.
This can also be seen in Fig.~\ref{img:augmented}.

During training, we insert in each position of the training minibatch a sample $(x_i, \mathbf{y}_i) \in \mathcal{D}$ with probability $\alpha$ or one of its synthetic variations $(x_{ij}, \mathbf{y}_i) \in \tilde{\mathcal{D}}$ with probability $(1-\alpha)$, where $x_{ij}$ is sampled uniformly from $S_i$. We use a value of $\alpha=0.5$, to balance real and generated images during training as suggested in \cite{trabucco}.

\section{Experiments}
In order to test our augmentation technique we perform multiple experiments over different tasks. As a first downstream task we train an image-captioning model using both augmented and non-augmented versions of the dataset. For this set of experiments, we selected medium-sized, Transformer-based Vision and Language models which can be trained end-to-end and can be employed for a variety of different tasks.  
In particular, we use the $\text{GIT}_\text{base}$ \cite{wang2022git} model and the $\text{BLIP}_\text{base}$ \cite{li2022blip} model.

GIT \cite{wang2022git} (Generative Image-to-text Transformer) is a Transformer \cite{vaswani2017attention} model which can be applied to many Vision and Language tasks. It leverages a CLIP ViT image encoder \cite{clip} and a single Transformer text decoder, which are jointly trained under a single language modeling task on large-scale pre-training data.
It is publicly available in two sizes, GIT-base (129 M parameters) which employs a CLIP/ViT-B/16 encoder and GIT-large (347M parameters), with a CLIP/ViT-L/14 encoder.
BLIP \cite{li2022blip} instead is a model that effectively uses noisy web data for pre-training by bootstrapping the captions, generating new synthetic captions and removing the noisy ones. It employs a multimodal mixture of encoder-decoder which are jointly trained with three vision-language objectives: image-text contrastive learning, image-text matching, and image-conditioned language modeling. The architecture is  composed of a ViT \cite{dosovitskiy2020image} encoder to encode images and a BERT \cite{devlin2018bert} to encode text.
Both the GIT and BLIP models were initialized with the available pre-training weights\footnote{https://huggingface.co/microsoft/git-base} and finetuned for 10 training epochs using the AdamW \cite{adamw} optimizer with a $5e^{-05}$ learning rate and 500 steps of warm-up using batches of 8 images. 

The second task is adopt to prove the effectiveness of our proposed strategy is cross-domain retrieval. Here, we perform retrieval both of images given their textual description and vice versa. For this downstream application, we use the CLIP model~\cite{clip}, using the openCLIP\footnote{https://github.com/mlfoundations/open\_clip} implementation.
To finetune the CLIP model we used again the AdamW optimizer with a learning rate of $5e^{-04}$.

\begin{table*}[t]
    \centering
    \begin{tabular}{l|c|c|c|c|c|c|c|c|c}
    Dataset & Model & B@1 & B@2 & B@3 & B@4 & METEOR & ROUGE & CIDEr & BERTScore\\
    \hline
    \multirow{6}{*}{\textbf{Artpedia}}
    &$\text{GIT}_b$ (zero-shot) & 0.0000 & 0.0000 & 0.0000 & 0.0000  & 0.0144 & 0.0749 & 0.0144& 0.6905\\
    &$\text{GIT}_b$ w/o DA & 0.0179 & 0.0088 & 0.0046 & 0.0026 & 0.0385 & 0.1433 & 0.0505& 0.7291\\
    &$\text{GIT}_b$ w/ DA & \textbf{0.0184} & \textbf{0.0092} & \textbf{0.0048} & \textbf{0.0027} & \textbf{0.0390} & \textbf{0.1479} & \textbf{0.0673} & \textbf{0.7316}\\
    \cline{2-10}
    &$\text{BLIP}_b$ (zero-shot) & 0.0000 & 0.0000 & 0.0000 & 0.0000 & 0.0161 & 0.0830 & 0.0043 & 0.7112\\
    &$\text{BLIP}_b$ w/o DA & 0.0050 & 0.0026 & 0.0014 & 0.0009 & 0.0331 & 0.1568 & 0.0766 & 0.7262 \\
    &$\text{BLIP}_b$ w/ DA & \textbf{0.0118} & \textbf{0.0062} & \textbf{0.0035} & \textbf{0.0020} & \textbf{0.0369} & \textbf{0.1658} & \textbf{0.0906} & \textbf{0.7291} \\
    \hline
    \hline
    \multirow{6}{*}{\textbf{ArtCap}}
    &$\text{GIT}_b$ (zero-shot) & 0.3993 & 0.2541 & 0.1548 & 0.0888 & 0.1237 & 0.3128 & 0.2114& 0.7877\\
    &$\text{GIT}_b$ w/o DA & 0.7311 & 0.5675 & 0.4263 & 0.3196 & 0.2360 & 0.5148 & 0.6263& \textbf{0.8752}\\
    &$\text{GIT}_b$ w/ DA & \textbf{0.7475} & \textbf{0.5825} & \textbf{0.4407} & \textbf{0.3321} & \textbf{0.2376} & \textbf{0.5166} & \textbf{0.6445} & 0.8737\\
    \cline{2-10}
    &$\text{BLIP}_b$ (zero-shot) & 0.6224 & 0.4007 & 0.2487 & 0.1512 & 0.1606 & 0.3951 & 0.3467 & 0.8098\\
    &$\text{BLIP}_b$ w/o DA & \textbf{0.7710}  & \textbf{0.5972} & 0.4515 & 0.3343 & 0.2442 & 0.5128 & 0.6851 & \textbf{0.8759}\\
    &$\text{BLIP}_b$ w/ DA & 0.7654 & 0.5909 & \textbf{0.4541} & \textbf{0.3491} & \textbf{0.2466} & \textbf{0.5170} & \textbf{0.6862} & 0.8748\\
    \end{tabular}
    \caption{Image Captioning results on Artpedia and ArtCap using GIT \cite{wang2022git} model, measuring BLEU ($n$-grams 1 to 4), METEOR, ROUGE, CIDEr and BERTScore metrics. }  
    \label{tab:n_gram1}
\end{table*}

\subsection{Quantitative Results}
\subsubsection{Metrics}
To quantitatively assess the quality of the generated captions, standard language evaluation metrics are used. Those include BLEU \cite{papineni-etal-2002-bleu}, ROUGE \cite{lin-2004-rouge} and METEOR \cite{denkowski2014meteor1_5}, typically used for machine translation tasks, and CIDEr \cite{vedantam2015cider}, specifically developed for the image captioning task. In addition, a semantic similarity between generated captions and references is measured with BERTScore \cite{zhang2019bertscore} metric.
BLEU score calculates $n$-gram precisions between a candidate sentence and a set of human-generated references, multiplied by a brevity penalty. Single $n$-gram precisions are then combined following a geometric mean to obtain a final score. It is common practice to report BLEU scores with $n$-grams ranging from 1 to 4. 
ROUGE-L calculates a F-measure using the Longest Common Subsequence (LCS) between a candidate sentence and a set of references. 
METEOR computes a harmonic mean of precision and recall between unigrams of aligned candidate and reference sentences, where the mapping used for alignment follows various strategies, including exact match, synonyms and paraphrases. 
CIDEr measures the consensus among a candidate sentence and a set of references by computing the cosine similarity of TF-IDF weighted $n$-gram vectors.
BERTScore uses the word embeddings computed by a pretrained Transformer model to measure the semantic similarity between a candidate sentence and a reference.

\subsubsection{Baselines}
\label{aug_meth}

\begin{figure}[t]
    \centering
    \begin{tabular}{ll}
    \rotatebox[origin=c]{90}{\scriptsize No Augmentation} &
    \includegraphics[width=.9\columnwidth,valign=m]{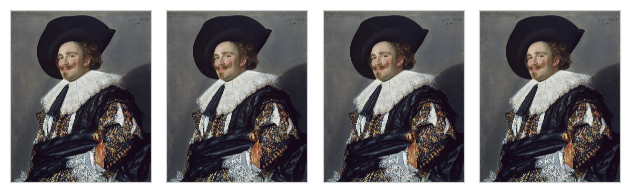}\\
    \rotatebox[origin=c]{90}{\scriptsize AutoAugment} &
    \includegraphics[width=0.9\columnwidth,valign=m]{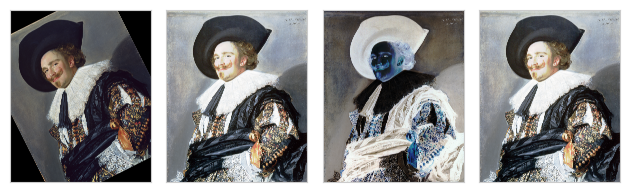}\\
    \rotatebox[origin=c]{90}{\scriptsize RandAugment} &
    \includegraphics[width=.9\columnwidth,valign=m]{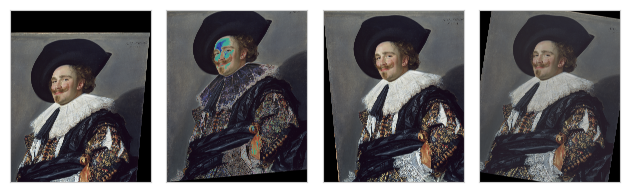}\\
    \rotatebox[origin=c]{90}{\scriptsize AugMix} &
    \includegraphics[width=.9\columnwidth,valign=m]{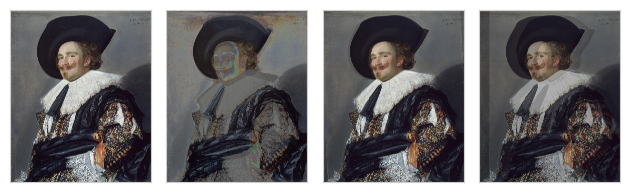}\\
    \rotatebox[origin=c]{90}{\scriptsize TrivialAugment} &
    \includegraphics[width=.9\columnwidth,valign=m]{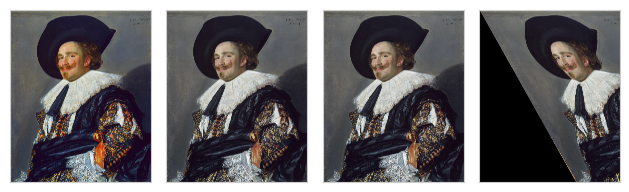}\\
    \rotatebox[origin=c]{90}{\scriptsize Ours} &
    \includegraphics[width=.9\columnwidth,valign=m]{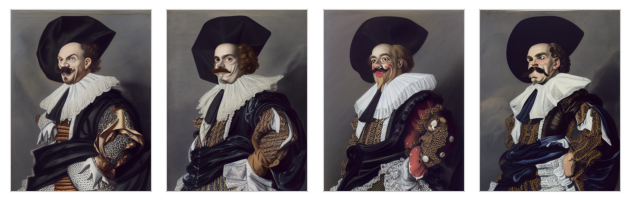}\\
    \end{tabular}

    \caption{Sample of images produced by different augmentation methods: No augmentation, AutoAugment \cite{autoaug}, RandAugment \cite{cubuk2020randaugment}, AugMix \cite{hendrycks2019augmix}, and TrivialAugment \cite{muller2021trivialaugment}, Ours. }
    \label{fig:augmentations}
\end{figure}

We compare our method against state-of-the-art augmentation techniques, such as AutoAugment \cite{autoaug}, AugMix \cite{hendrycks2019augmix}, RandAugment \cite{cubuk2020randaugment} and TrivialAugment \cite{muller2021trivialaugment}.
AutoAugment \cite{autoaug} is an augmentation framework for vision models that casts the search of parameters for data augmentation as an optimization problem and solves it using reinforcement learning. RandAugment \cite{cubuk2020randaugment} improves on AutoAugment \cite{autoaug} by both considerably reducing the parameters search space from $10^{32}$ to $10^{2}$ and matching or exceeding performances of \cite{autoaug}. AugMix \cite{hendrycks2019augmix} layers multiple randomly sampled augmentation operations in concert with a consistency loss to improve model robustness. Finally \cite{Muller_2021_ICCV} improves on the previous strategies by further simplifying the search space.
All of the previous models are tailored to image classification tasks, more recently a number of works focused on data augmentation specifically developed for detection problems. In Fig.~\ref{fig:augmentations} we provide a comparison of the augmentation operations performed by the aforementioned state of the art techniques and ours applied to the same image.

\begin{table*}[t]
    \centering
    \begin{tabular}{l|c|c|c|c|c|c}
             & No aug & AutoAugment & RandAugment & AugMix 
             & TrivialAugment & Ours \\ 
    \hline
    Artpedia & 0.0505 & 0.0583 & 0.0533 & 0.0536 
            & 0.0510 & \textbf{0.0673}    \\ 
    ArtCap   & 0.6263 & 0.5829 & 0.6239 & 0.5717 
    & 0.5849 & \textbf{0.6445}    \\ 
    \end{tabular}
    \caption{Comparison of CIDEr scores with GIT \cite{wang2022git} model trained using our proposed diffusion augmentation and other state-of-the-art augmentation techniques for the image captioning task: AutoAugment \cite{autoaug}, RandAugment \cite{cubuk2020randaugment}, AugMix \cite{hendrycks2019augmix}, and TrivialAugment \cite{muller2021trivialaugment}.}
    \label{tab:augmentations}
\end{table*}

\subsubsection{Results}
\paragraph{Image Captioning} We present the results of the image captioning task using differently trained GIT models in Tab.~\ref{tab:n_gram1} over the two chosen datasets.
For the Artpedia dataset, the test results show a clear and consistent improvement using our augmentation technique over all the aforementioned metrics.
Similarly, on ArtCap we report gains in most metrics, with only a slight decrease compared to standard training with no data augmentation.
It is also easy to notice how the two datasets differ in terms of complexity. In fact, all the models struggle more with Artpedia~\cite{stefanini2019artpedia}, obtaining results that are much lower in absolute terms. This is due to the nature of the captions in Artpedia, which are composed of long sentences, with lots of details, as can be seen in Fig. \ref{fig:dataset_samples} and Fig. \ref{img:augmented}. Therefore, n-gram-based metrics fail to effectively convey the quality of the captions. On the contrary, BERTSCore, which captures semantic similarity between sentences rather than analyzing them from a structural point of view, achieves much higher results and confirms the improved quality of the captions generated with our data augmentation.
On the contrary, ArtCap has shorter sentences so metrics such as BLEU, METEOR, ROUGE and CIDEr manage to obtain much higher results in absolute terms and can be used effectively in this comparison.

In order to assess the quality of our data augmentation strategy we also present a comparison between our method and different state of the art methods for image data augmentation (from Section \ref{aug_meth}) in Tab.~\ref{tab:augmentations}.
It is important to note that while our augmentation approach is beneficial to the model, other augmentation techniques actually hurt performance. Intuitively we can infer that data augmentation strategies such as the one we compare our method against are engineered for classification tasks and might not might semantically invariant with reference to image captioning.

\paragraph{Image Retrieval} For the retrieval task we test CLIP \cite{clip} using a similar setting to image captioning. We test first in a zero-shot configuration and then with and without data augmentation.
The CLIP model is pre-trained on the YFCC dataset \cite{thomee2016yfcc100m} when performing zero-shot retrieval. As in the previous task, the results shown in Tab.~\ref{tab:clip} present a clear indication of an improved performance in the retrieval problem.
Tab.~\ref{tab:comparison_artpedia} instead compares our results with the best ones proposed in \cite{stefanini2019artpedia} using the same experimental protocol by the authors, i.e. by fixing the maximum number of retrievable items to $N=100$.

\begin{table}[]
    \centering
    \begin{tabular}{l|c|c|c|c}
    Model & Task & R@1 & R@5 & R@10\\
    \hline
         CLIP - (zero-shot) & im2t & 0.0853 & 0.1557 & 0.2096  \\
         CLIP - w/o DA & im2t & 0.1048 & 0.2081 &0.2665  \\
         CLIP - w/ DA & im2t & \textbf{0.1108} & \textbf{0.2096} & \textbf{0.2740}  \\
         \hline
         CLIP - (zero-shot) & t2im & 0.0644& 0.1751& 0.2290  \\
         CLIP - w/o DA & t2im & \textbf{0.0883} & 0.1751 & 0.2305  \\
         CLIP - w/ DA & t2im & 0.0868 & \textbf{0.1976} & \textbf{0.2470}  \\
    \end{tabular}
    \caption{Test on Artpedia on the retrieval task with CLIP using a ResNet50 pretrained on YFCC \cite{thomee2016yfcc100m}. We report Recall~@1,@5,@10. We test both in the image-to-text (im2t) setting and in the text-to-image (t2im) setting.}
    \label{tab:clip}
\end{table}

\begin{table}[]
    \centering
    \begin{tabular}{l|c|c|c}
    Model & Task & R@1 & R@5 \\
    \hline
         X-Attn GloVe \cite{stefanini2019artpedia} & im2t & 0.086 & 0.227 \\
         CLIP - w/ DA & im2t & \textbf{0.090}	& \textbf{0.230}  \\
         
         \hline
         X-Attn GloVe \cite{stefanini2019artpedia} & t2im & 0.041 & 0.136 \\
         CLIP - w/ DA & t2im & \textbf{0.090}	& \textbf{0.250}  \\
         
    \end{tabular}
    \caption{We compare our results using the same experimental protocol as in \cite{stefanini2019artpedia} using $N=100$ retrievable items. We report Recall~@1 and @5, testing both in the image-to-text (im2t) setting and in the text-to-image (t2im) setting. }
    \label{tab:comparison_artpedia}
\end{table}

\subsection{Qualitative Results}

\begin{figure}
    \centering
    \includegraphics[width=0.98\columnwidth]{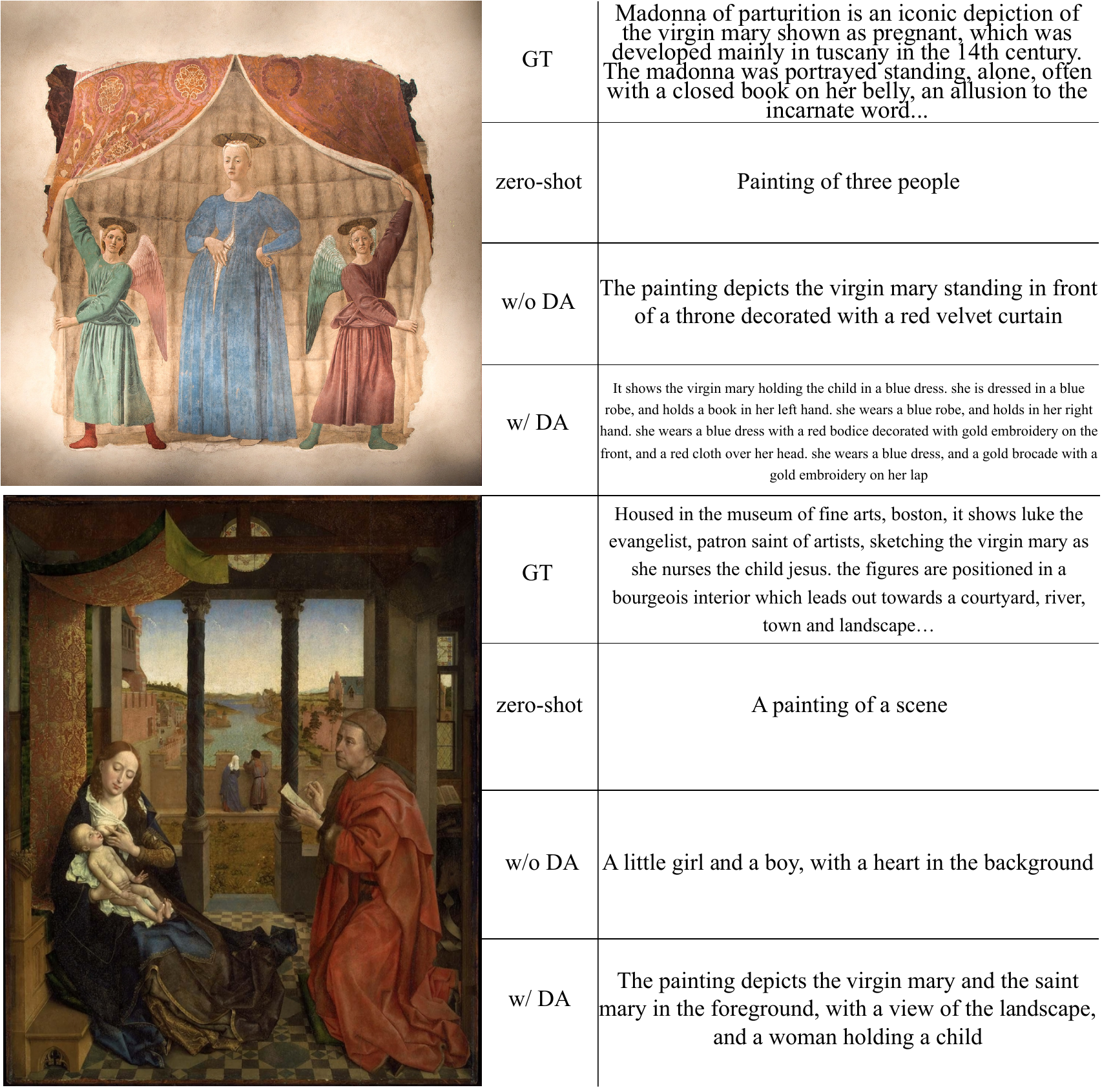}
    \caption{Qualitative samples showing the original (GT) caption along with different outputs from GIT \cite{wang2022git} on Artpedia \cite{stefanini2019artpedia} }
    \label{fig:luke}
\end{figure}

Due to the subjective nature of the task, it is necessary to perform a visual inspection to better understand how are the models behaving under different data regime conditions.
In this section, we present a sample of qualitative results to better appreciate the effect of our method. The example presented in Fig.~\ref{fig:luke} shows the effectiveness of our model in terms of enriching the quality of the description by comparing the output of the same captioning method trained under different settings.

While the pre-trained model tends to offer vague but correct descriptions even in a zero-shot setting, it is necessary to fine-tune the model on the target dataset in order to match the language used in Artpedia dataset. Our data-augmented finetuning helps the model to obtain a better representation of fine visual details in the dataset, allowing to obtain richer captions using the task-related technical knowledge that a large internet-wide trained model might be missing.

\section{Conclusions}
This paper presented technique for augmenting and better exploit fine art datasets with the intent of making the fruition of semantically complex visual art easier to digitalise, to access, and to retrieve for the general public. In the field of cultural heritage a feature such as the uniqueness of the artworks can become an obstacle for machine learning techniques that requires large amount of data. At the same time the usual augmentation techniques such as image flipping, random brightness change, random hue change do not suit the task as they semantically change the original datapoint by changing small the visual details that are actually meaningful.
Therefore our contributions aims at semantically enrich the popular pretrained LLMs models leveraging the expert knowledge to create a more sophisticated image data augmentation pipeline.

\section*{Acknowledgements}
This work is partially supported by the European Commission under European Horizon 2020 Programme, Grant No. 101004545-ReInHerit.

{\small
\bibliographystyle{ieee_fullname}
\bibliography{egbib}
}

\end{document}